\documentclass[preprint,12pt]{elsarticle}
\usepackage[margin=2.5cm]{geometry}
\usepackage{amssymb}
\journal{arXiv}

\usepackage{graphicx}
\usepackage{adjustbox}
\usepackage{amsmath}
\usepackage{amsthm}
\newtheorem{definition}{Definition}
\usepackage{amssymb}
\usepackage{tabularx}
\usepackage{newtxmath}
\usepackage{mathbbol}
\usepackage{lineno}
\usepackage[hidelinks]{hyperref}
\usepackage{soul}

\usepackage[dvipsnames]{xcolor}
\usepackage{subfiles}
\usepackage{tikz}
\usepackage{pgfplots}
\usepgfplotslibrary{colorbrewer}
\pgfplotsset{compat=1.18}
\usetikzlibrary{positioning, calc, backgrounds, shapes.geometric, shapes.arrows, arrows.meta, shapes.symbols, decorations.text, fit}
\usetikzlibrary{external}
\usepgfplotslibrary{colormaps}
\usepgfplotslibrary{statistics}
\usepackage[most]{tcolorbox}
\usepackage{hyperref}

\begin{document}

\begin{frontmatter}

\title{Fully data-driven inverse hyperelasticity with hyper-network neural ODE fields}
\author[inst1]{Vahidullah Taç\corref{contrib}}
\author[inst1]{Amirhossein Amiri-Hezaveh\corref{contrib}}
\author[inst2]{Manuel K. Rausch}
\author[inst2]{Grace N. Bechtel}
\author[inst3]{Francisco Sahli Costabal}
            \ead{corresponding authors: fsc@ing.puc.cl}
\author[inst1,inst4]{Adrian Buganza Tepole}
            \ead{abuganza@purdue.edu}
\affiliation[inst1]{organization={Department of Mechanical Engineering, Purdue University}, 
            city={West Lafayette},
            state={IN},
            country={USA}}
\affiliation[inst2]{organization={Department of Aerospace Engineering and Engineering Mechanics, The University of Texas at Austin}, 
            city={Austin},
            state={TX},
            country={USA}}
\affiliation[inst3]{organization={Department of Mechanical and Metallurgical Engineering and Institute for Biological and Medical Engineering, Pontificia Universidad Catolica de Chile}, 
            city={Santiago},
            country={Chile}}

\affiliation[inst4]{organization={Weldon School of Biomedical Engineering, Purdue University}, 
            city={West Lafayette},
            state={IN},
            country={USA}}
\cortext[contrib]{Authors contributed equally}
\begin{abstract}
We propose a new framework for identifying mechanical properties of heterogeneous materials without a closed-form constitutive equation.  Given a full-field measurement of the displacement field, for instance as obtained from digital image correlation (DIC), a continuous approximation of the strain field is obtained by training a neural network that incorporates Fourier features to effectively capture sharp gradients in the data. A physics-based data-driven method built upon ordinary neural differential equations (NODEs) is employed to discover constitutive equations. The NODE framework can represent arbitrary materials while satisfying constraints in the theory of constitutive equations by default. To account for heterogeneity, a hyper-network is defined, where the input is the material coordinate system, and the output is the NODE-based constitutive equation. The parameters of the hyper-network are optimized by minimizing a multi-objective loss function that includes penalty terms for violations of the strong form of the equilibrium equations of elasticity and the associated Neumann boundary conditions. We showcase the framework with several numerical examples, including heterogeneity arising from variations in material parameters, spatial transitions from isotropy to anisotropy, material identification in the presence of noise, and, ultimately, application to experimental data.  As the numerical results suggest, the proposed approach is robust and general in identifying the mechanical properties of heterogeneous materials with very few assumptions, making it a suitable alternative to classical inverse methods.             
% Heterogeneous materials ubiquitous in nature, tissues incorporate different structures for multi-function, and heterogeneities can be sign of disease for example stiff tumor inclusions in soft tissues. Engineered materials can leverage heterogeneity to produce materials with extreme mechanical properties, and heterogeneous properties can also be used to monitor structural health. With progress in machine learning and AI, able to generate heterogeneous materials for optimal design. The inverse problem has been explored for simple closed form materials such as neo-Hookean. Yet, many real materials exhibit highly nonlinear mechanics with no adequate close form model. In previous work we have developed fully data-driven constitutive modeling and genrative AI for exploring heterogeneous data-driven materials. Here we solve the inverse problem of finding data-driven materials in the form of neural ODEs (NODE) fields, without any assumption about the underlying material behavior except polyconvex hyperelasticity. The method relies on a hyper-network approach which produces the NODE field and evaluates mechanical equilibrium given a set of boundary conditions and a full-field strain field. The method outperforms standard techniques such as the virtual fields method, and can learn complex geometries and works accurately with experimental data.  
\end{abstract}

\begin{keyword}
Hyperelasticity \sep inverse problems \sep neural ODEs \sep data-driven constitutive model \sep heterogeneous materials \sep hyper-networks
\end{keyword}

\end{frontmatter}

%% main text
\section{Introduction}

% why do we care? when a material is heterogeneous it is difficult to get properties of the different parts of the material in physiological test conditions
Heterogeneous materials are ubiquitous in both engineered systems and biological tissues, and accurately characterizing their mechanical behavior is essential for design, performance prediction, and diagnostic insight. For example, composites, a significant class of heterogeneous materials, have found many applications in mechanical, aerospace, automotive, and civil engineering, to name a few \cite{sharma2022advances}. In the human body, tissues often exhibit heterogeneous properties by packing distinct cell types and materials to satisfy various physiological demands \cite{lin2023impact,sack2023magnetic,moreno2024role}. 
At the same time, heterogeneity can be a sign of disease. For instance, breast cancer is characterized by stiff inclusions in soft tissue \cite{koorman2022spatial}, and it is often diagnosed by palpating the stiff inclusion in the softer breast tissue \cite{olson2019inverse}. Hence, the identification of heterogeneity in mechanical properties is also essential to understanding the fundamental mechanics of tissues and to improving the diagnosis of various diseases \cite{chai2015local,zhang2021finite}.  

% introduced examples of heterogeneous structures, but havent said why it is actually difficult to get heterogeneous properties 
A common strategy for probing spatial heterogeneity in mechanical properties involves localized testing techniques such as nanoindentation \cite{qian2018nanoindentation} and atomic force microscopy (AFM) \cite{azeloglu2011atomic}. While these methods offer high spatial resolution, they are technically demanding to perform reliably and often yield measurements that are not representative of physiological loading conditions. In particular, many soft biological tissues such as heart valves, tendons, and skin, experience large, anisotropic tensile deformations in vivo, whereas indentation tests primarily impose localized compressive loads in the small deformation regime \cite{pailler2008vivo,moghaddam2020indentation,cox2006mechanical}. This mismatch in loading modality can result in mechanical properties that are not predictive of functional behavior under physiologically relevant conditions \cite{kingsley2019multi}. As a result, there is increasing interest in alternative techniques that provide spatially resolved, physiologically meaningful mechanical characterization \cite{luetkemeyer2021constitutive}.
%One way to avoid such challenges is to develop a computational framework that leverages the physics law while using feasible experimental data as input to identify material properties---an approach known as inverse methods. 

% Ok, at least now I said that local properties are different from physiology. And that was a good way of introducing the need for inverse problems 
With the advent of high-resolution imaging and advanced image processing, full-field experimental methods such as digital image correlation (DIC) have become widely adopted for characterizing mechanical behavior \cite{rastogi2003photomechanics,claire2004finite,avril2008overview}. These techniques enable the capture of rich spatially resolved displacement and strain fields under realistic loading conditions. For soft materials and thin structures, 2D and stereo DIC are routinely used, while volumetric imaging modalities such as MRI, CT, and ultrasound are increasingly employed for 3D biological tissues \cite{genovese2021multi,estrada2020mr}. Compared to pointwise or localized mechanical tests, full-field approaches offer a more physiologically relevant picture of material response and eliminate the need to repeat spatially localized experiments that often require simplifying assumptions.

To extract constitutive information from these rich datasets, several inverse modeling techniques have been developed. These include the virtual fields method (VFM) \cite{toussaint2006virtual,grediac2006applying,pierron2012virtual}, finite element model updating (FEMU) \cite{kavanagh1971finite,avril2008overview,pottier2011contribution}, the constitutive equation gap method (CEGM) \cite{geymonat2002identification,florentin2010identification}, the reciprocity gap method \cite{ikehata1990inversion}, and the equilibrium gap method \cite{claire2004finite,claire2007identification}. These approaches formulate the identification task as a constrained optimization problem, relying on a known constitutive model and enforcing balance laws to infer spatial variations in material parameters from the measured fields.

Most classical inverse methods for material characterization mentioned above are based on the weak form of the equilibrium equations. This formulation is advantageous for standard finite element discretizations because it reduces regularity requirements. However, in techniques such as FEMU,  repeated solution of the forward problem as well as the computation of adjoint sensitivities is needed , making the process computationally intensive \cite{kavanagh1971finite,avril2008overview,pottier2011contribution}. In VFM, test functions (termed virtual fields) must be chosen carefully to isolate specific parameters, a step that is often problem-dependent and lacks a general prescription \cite{toussaint2006virtual,grediac2006applying,pierron2012virtual}. More broadly, all weak-form approaches require integration over the domain, which introduces sensitivity to quadrature schemes, mesh resolution, and the spatial support of test functions. In contrast, strong-form formulations enforce equilibrium directly at collocation points and avoid the need for integration or test function design. With the advent of smooth neural interpolants and automatic differentiation, enforcing the strong form has become not only feasible but also computationally efficient and highly scalable in  physics-informed machine learning (ML) \cite{karniadakis2021physics}.

Arguably, the biggest limitation of existing inverse identification of material properties is that they are limited to closed-form constitutive models and often impose strong restrictions on the geometry of the heterogeneities. Data-driven constitutive modeling has gained traction by leveraging the universal approximation capabilities of neural networks to represent arbitrary stress-strain behavior without relying on a predefined functional form \cite{ghaboussi1991knowledge,as2022mechanics}. However, this flexibility introduces new challenges: ensuring physical plausibility, enforcing material constraints, and performing robust inverse identification under noisy and partial observations. Recent works have addressed these limitations by incorporating thermodynamic constraints, objectivity, and normality conditions either via penalty methods or architectural design \cite{fuhg2022physics,linden2023neural,tacBenchmarkingPhysicsinformedFrameworks2023a}. Of particular interest are recent developments using neural ordinary differential equations (NODEs) \cite{tac2022node}, which can naturally satisfy key constitutive constraints—such as energy positivity, stress vanishing in the reference configuration, and polyconvexity—without requiring ad hoc regularization.

In this work, we present a new framework for inverse identification of heterogeneous materials using full-field  data, building on the NODE-based constitutive models introduced in \cite{tac2022node}. Unlike traditional methods that assume a closed-form constitutive law and solve a parametric optimization problem, our approach is fully data-driven, with material behavior learned directly from the data. To accommodate spatial variation, we introduce a hyper-network that maps each material point to a set of NODE parameters, thereby enabling arbitrary heterogeneous material response fields \cite{kirchhoff2024inference}. In other words, we directly work to learn the infinite dimensional material fields. The result is a robust and flexible inverse method that respects physics constraints by construction and scales naturally to complex, heterogeneous materials.

\begin{figure}[h!]
    \centering
    % \subfile{figures/fig_architecture}
    \begin{adjustbox}{center}
        \includegraphics[width=1.2\textwidth]{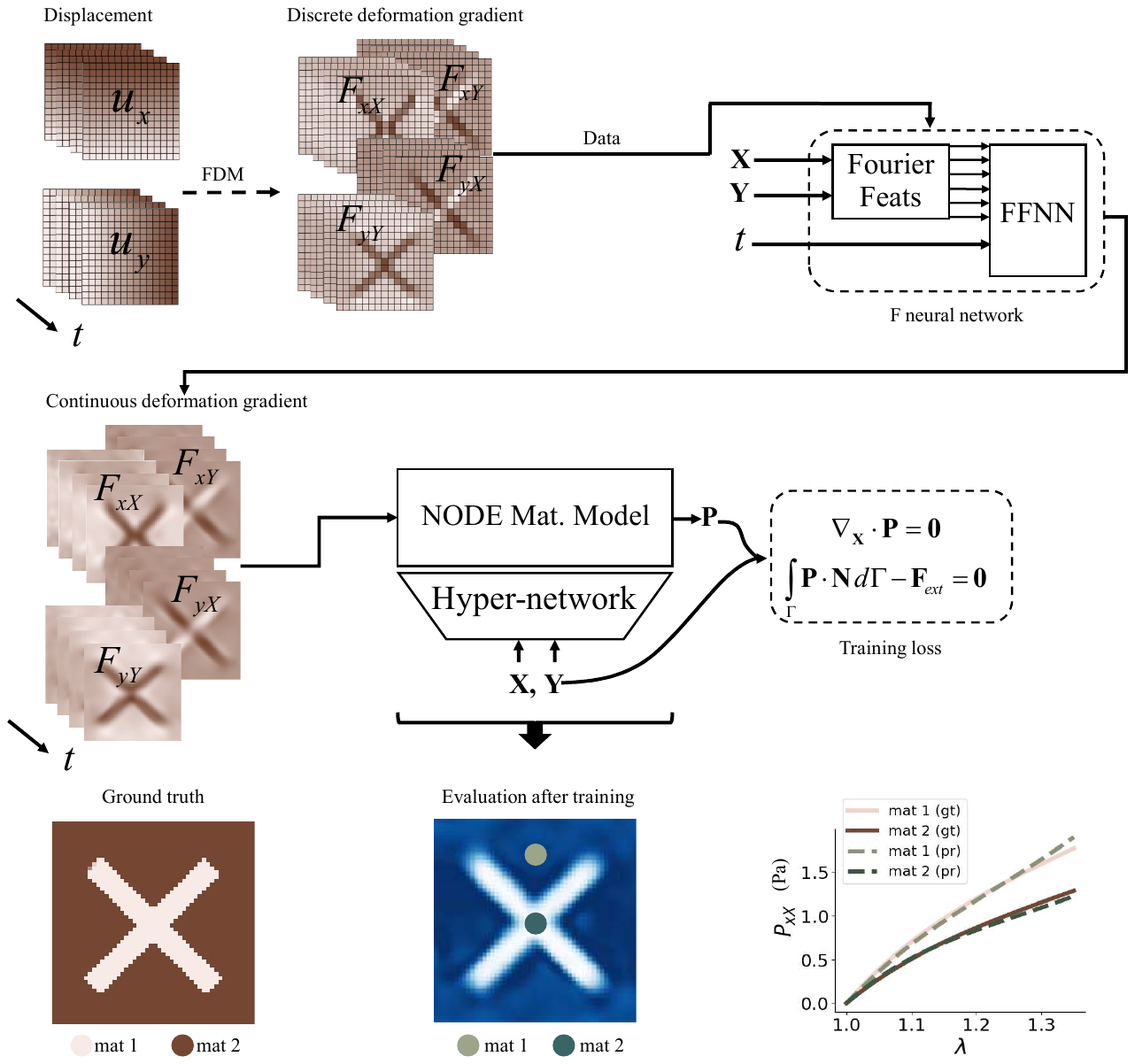}
    \end{adjustbox}
    
    \caption{Detailed overview of the data-driven inverse method. The framework begins with raw displacement field data, which can be obtained from experimental observations or, here, through a numerical example. A finite difference scheme is used to compute discrete deformation gradients. These tensorial quantities are interpolated using a neural network augmented with Fourier features to capture sharp spatial variations with a continuous approximation. To characterize the underlying constitutive behavior, a NODE-based model is employed. In this formulation, a hyper-neural network defines the spatially dependent parameters of the NODEs, and the entire model is trained using a loss function that enforces both the strong form of the equilibrium equations and traction boundary conditions.}  
    \label{fig_summary} 
\end{figure}

\begin{figure}[h!]
    \centering
    % \subfile{figures/fig_summary}
    \begin{adjustbox}{center}
        \includegraphics[width=1.2\textwidth]{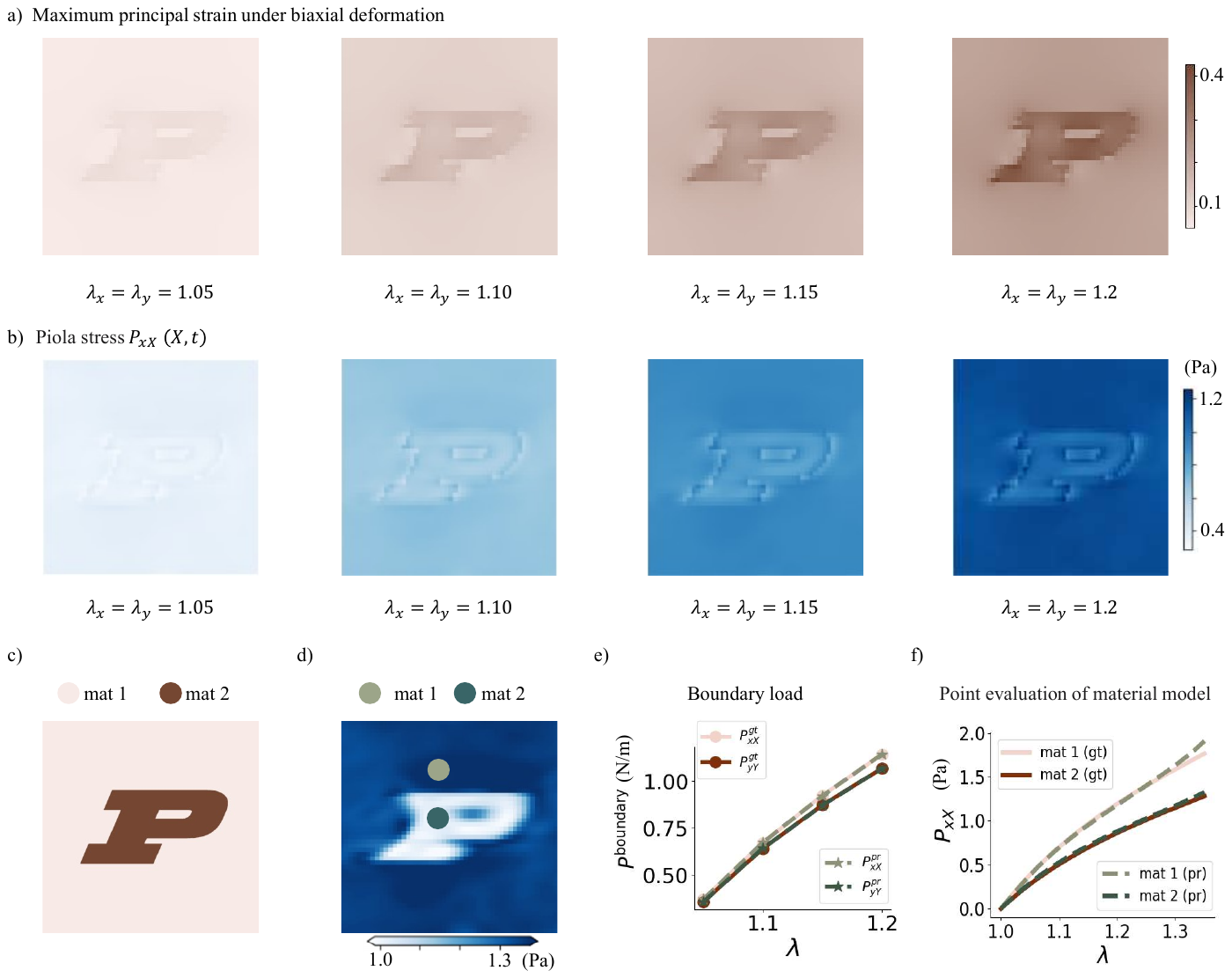}
    \end{adjustbox}
    
    \caption{Identification of inhomogeneity through the proposed data-driven inverse method. A soft "P"-shaped inclusion is embedded within a stiff matrix, with both phases modeled as neo-hookean materials with different material parameters. (a) Ground truth maximum principal strain field under biaxial loading. (b) Nominal stress distribution obtained after training, based on the corresponding equilibrium strain data. (c) Ground truth and (d) identified material modeled represented as the trace of the stress at a spatially-uniform biaxial deformation.  (e) Boundary force distributions and (f)  stress–strain responses at two selected points, comparing predictions against the ground truth. }
    \label{p_example}
\end{figure}

% npj computational materials puts results first 
\section{Results}

\subsection{Recovering arbitrary heterogeneous material fields from strain field data and boundary conditions}

% Fig 2. Details of the method as a 'result' 
The details of the new solution procedure are  illustrated in Fig \ref{fig_summary}. Note that the goal is to obtain a Piola stress field $\bf{P}(\bf{X})$ that  satisfies equilibrium, i.e., $\nabla_{\bf{X}} \cdot {\bf{P}} = \mathbf{0}$ in the domain $\Omega$, along with the prescribed traction boundary conditions. As we explain in the Methodology (Section~\ref{method_sec}), to find such a stress field, we minimize a loss function containing both terms (referred to as  \textit{Training loss} in Fig \ref{fig_summary}). The stress field, in turn, is evaluated from ${\bf{P}}=2{\bf{F}}\,\,\partial \Psi(\bf{F})/\partial \bf{C}$, in which the strain density function $\Psi$ is approximated in terms of summation of NODEs (shown as \textit{NODE Mat. Model} in Fig \ref{fig_summary}, see also \eqref{eq_strain energy density function_general}), where the deformation gradient field $\bf{F}$  is either directly provided as input data or calculated from the displacement field data (Fig \ref{fig_summary}).  Because of the divergence operator in the loss function, we need a spatially continuous representation of $\bf{F}$. In this regard, there are two options: interpolating $\bf{u}$ continuously and differentiating to get $\bf{F}$. Alternatively, one can apply a finite difference method, such as the central difference method, on discrete $\bf{u}$, and subsequently interpolate $\bf{F}$ continuously from the discrete field with a neural network. In Fig \ref{fig_summary}, We assume that displacements are given as the input data. Next, using finite differences we obtain the discretized deformation gradient field. Subsequently, a continuous interpolation of the discrete $\bf{F}$  is obtained from a feed-forward neural network with Fourier features encoding to capture the sharp gradients. The key component of the method is the representation of the strain energy function field $\Psi({\bf{X}},{\bf{F}},\rm\,{cof}{\bf{F}},J)$. We recover a full material model point-wise. In other words, for each $\bf{X}$, we recover a polyconvex constitutive equation $\Psi_\Theta(\bf{F},\mathrm{cof}\bf{F},\det\bf{F})$ in terms of summation of NODEs as in our previous work\cite{tac2022node}. The notation $\Theta$ is used for the parameters of the NODE model. To discover the heterogeneous material response, the field of parameters $\Theta(\bf{X})$ is obtained from a multi-layer perceptron (MLP), referred to as hyper-network in Fig \ref{fig_summary}, with learnable parameters $\theta^{\text{hyper}}$. Then, the outcomes are the identified profile of the inhomogeneity and the corresponding stress–strain relations, as shown in Fig \ref{fig_summary} for a square with an  \textit{X}-shaped inhomogeneity. In particular, because the material field is parameterized by $\Theta(\bf{X})$ and it is not a closed-form model, we report the trace of the stress corresponding to a uniform deformation $\lambda_x(\mathbf{X}) = \lambda_y(\mathbf{X}) = 1.1$, as depicted in Fig \ref{fig_summary} by the blue square with a white \textit{X}. Also, for two selected points, the stress-stretch response under biaxial tension is shown in Fig \ref{fig_summary}, showing excellent agreement between ground truth and the NODE model at those locations. To obtain the above results, three sets of function approximators were employed. The interpolation MLP has an architecture of $[81, 40, 40, 4]$, with an initial Fourier feature layer of size $[2, 40]$. The NODEs use core MLPs with architecture $[1, 4, 4, 4, 1]$. The architecture of the MPL hyper-network is $[80, 40, 40, 126]$, with an initial Fourier feature layer of size $[2, 40]$.

% Fig 1 Show the pipeline with a result as an excuse. We get Psi from F
We further test out the methodology with a synthetic plate with a $P$-shape inhomogeneity in the middle, as shown in Fig \ref{p_example}. The true material response can be found in  Fig \ref{p_example}f. In this case, the inclusion and the surrounding materials are both neo-Hookean with shear moduli $\mu=2,1$ Pa, respectively. The input data consists of deformation gradient field  computed using JAX-FEM \cite{jaxFEM} at selected increments of equi-biaxial loading (see Fig \ref{p_example}a), along with traction boundary conditions. The outcomes of the inverse problem are the identification of the stress field ${\bf{P}}({\bf{X}},t)$ (Fig \ref{p_example}b) and the characterization of the underlying material response (Figs. \ref{p_example}e and f). While not explicitly shown, the identified stress field  ${\bf{P}}({\bf{X}},t)$ in Fig \ref{p_example}b  satisfies the equilibrium condition $\nabla_{\bf X} \cdot {\bf P}({{\mathbf {X}}}, t) = 0$, which is enforced by minimizing the loss function during the optimization of the hyper-neural network parameters. Satisfaction of boundary load data is shown in Fig \ref{p_example}e. The identified material is not closed form and, as a result, cannot be represented in terms of a heterogeneous modulus field. Rather, Fig \ref{p_example}d shows the trace of the stress for a uniform deformation $\lambda_x({\bf X})=\lambda_y({\bf X})= 1.1$. Note that the stress in Fig \ref{p_example}e is not an equilibrium stress field. Instead, it should be interpreted as an approximation to a modulus field. This is an approximation; in reality, at every point $\bf{X}$, we have a polyconvex strain energy potential, $\Psi({\bf{X}},{\bf{F}},\rm\,{cof}{\bf{F}},J)$. Fig \ref{p_example}f shows the evaluation of the resulting Piola stress from the polyconvex energy for two points, as well as comparison against the true neo-Hookean models. Thus, the proposed algorithm is able to capture an equilibrium stress field and a  field of data-driven NODE strain energy potentials, given full-field strains and boundary force data. To generate the above results, an MLP with architecture $[201, 40, 40, 4]$ and an initial Fourier feature layer of size $[2, 100]$ was used to interpolate the deformation gradient field. All NODEs share the same architecture as above: $[1, 4, 4, 4, 1]$. Additionally, the hyper-network MLP has architecture $[80, 40, 40, 126]$, with an initial Fourier feature layer of size $[2, 40]$ as above.

\begin{figure}[h!]
    \centering
    % \subfile{figures/fig_architecture}
    \begin{adjustbox}{center}
        \includegraphics[width=1.2\textwidth]{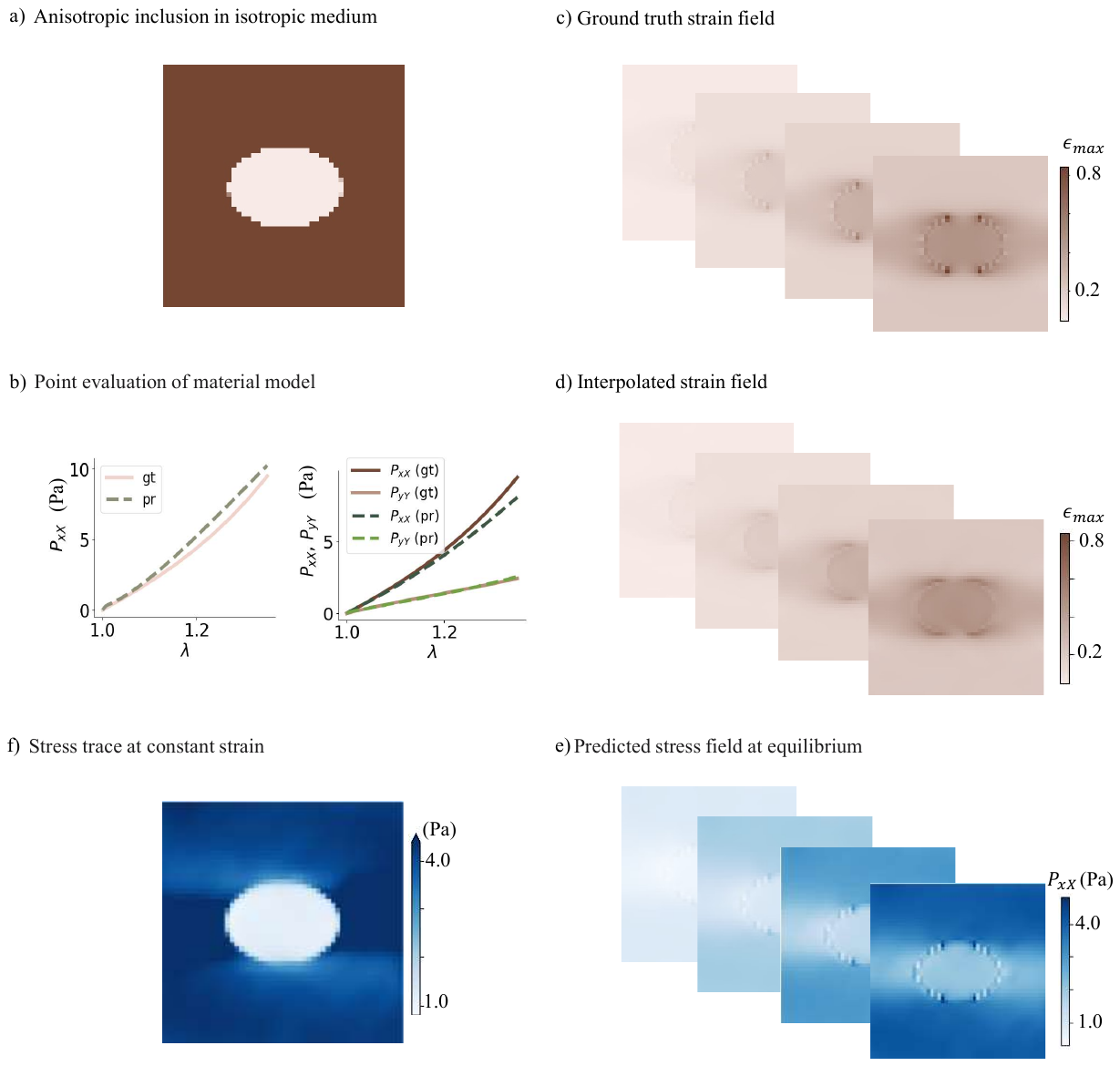}
    \end{adjustbox}
    \caption{Discovering material anisotropy. A  GOH material elliptical inclusion is embedded in a neo-Hookean material (a). (b) Predicted stress–strain responses for both isotropic and anisotropic materials for two spatial points in the continuum body. (c) Ground truth data and (d) the corresponding interpolated deformation gradient field, respectively. (e) Stress field predicted by the model in response to the ground truth equilibrium deformations.   }
    \label{anisotropy_example} 
\end{figure}

\subsection{Discovering anisotropy}

% Fig 3 
One of the main features of the proposed method is that there is no restriction on the material field except continuity by the nature of $\Theta(\bf{X})$ approximated by the MLP hyper-network. However, there is no specific format existing in closed form, and there is no assumption on the type of material or type of heterogeneity. In this regard, in the synthetic example of Fig \ref{anisotropy_example}a we select one region to be isotropic neo-Hookean, and the another region is anisotropic and based on the soft tissue constitutive model proposed by Gasser, Ogden,  (GOH) \cite{gasser2005GOH}. The sample was subjected to biaxial loading in JAX-FEM, and we report the discretized strain field directly from the finite element simulation in Fig \ref{anisotropy_example}c, and the interpolated strains in Fig \ref{anisotropy_example}d  using the Fourier feature neural network. The loss, as before, consists of linear momentum balance in the static limit and boundary loads. The equilibrium stress field after training is shown in Fig \ref{anisotropy_example}e. The trace of the stress at a constant strain field is depicted in Fig \ref{anisotropy_example}f to compare against the ground truth material in Fig \ref{anisotropy_example}a. Note that the model automatically discovers the anisotropy. Evaluation of the ground truth material behavior against the discovered NODE model at two locations is shown in  Fig \ref{anisotropy_example}b, confirming that the training naturally leads to anisotropic response prediction in the inclusion, compared to the isotropic response in the surrounding region. In this example, the following architectures were employed: $[201, 40, 40, 4]$ with an initial Fourier feature layer of size $[2, 100]$ for interpolating the ground truth deformation gradient field $\mathbf{F}$; $[1, 4, 4, 4, 1]$ for the MLP cores used in the NODEs; and $[20, 40, 40, 126]$ with an initial Fourier feature layer of size $[2, 10]$ for the MLP hyper-network.   
\begin{figure}[h!]
    \centering
    % \subfile{figures/fig_architecture}
    \begin{adjustbox}{center}
        \includegraphics[width=1.1\textwidth]{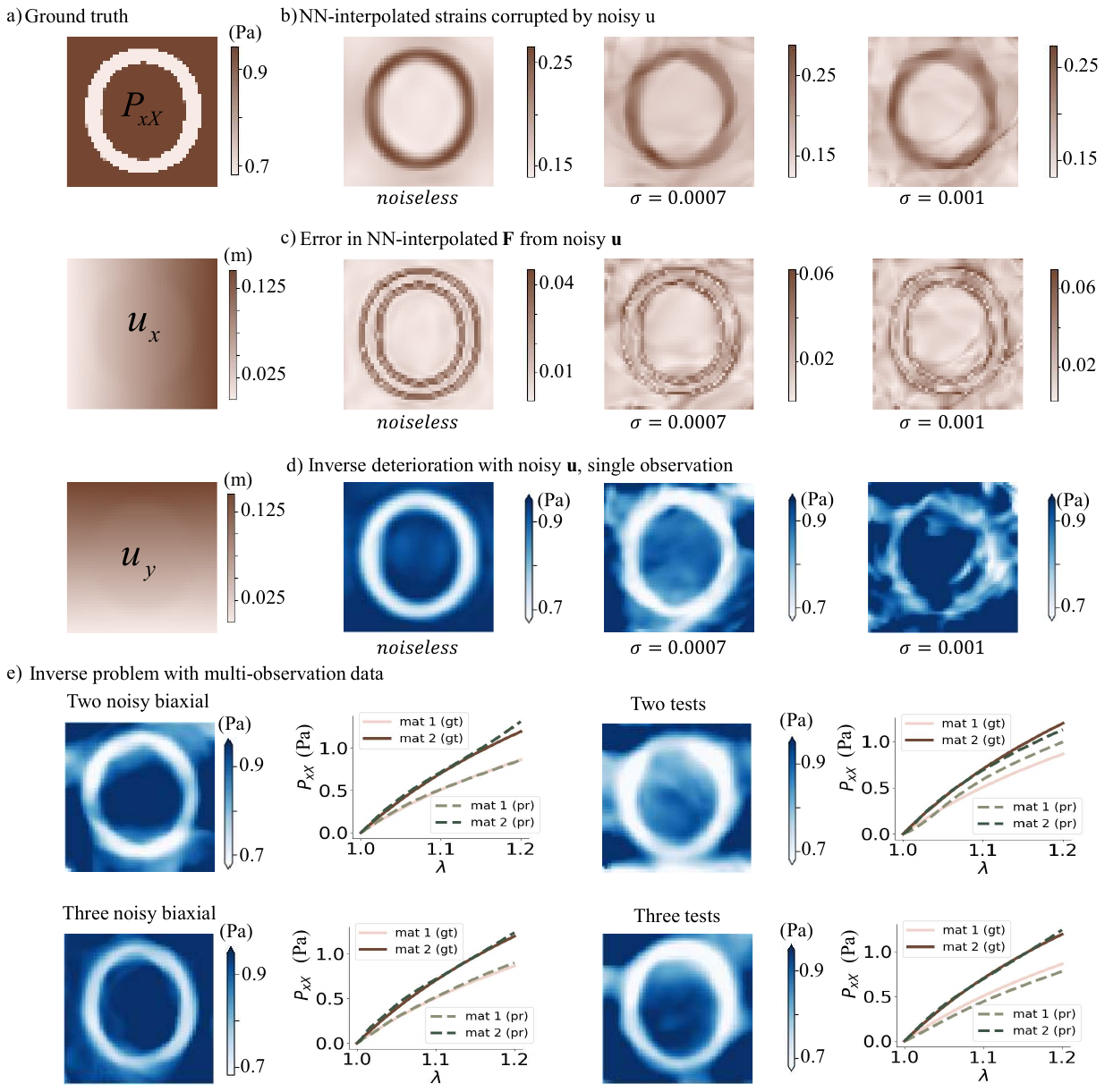}
    \end{adjustbox}
    
    \caption{Effect of noise. 
A ring-shape inclusion is embedded within a square matrix domain, with both regions composed of GOH materials with different parameter sets. To create noisy data, an artifact standard normal distribution noise is added to the ideal, noiseless ground truth displacement field (a). 
(b) Maximum principal strain fields for the noiseless case and for two noisy datasets. The corresponding norm of the error between interpolated deformation gradient fields and the ground truth ones are shown in  (c).  (d) Failure of the inverse method to accurately characterize the material when only a single noisy dataset is used.  (e) Recovery of the inverse characterization from noisy data with $\sigma = 0.0007$, achieved by increasing the amount of data, either through repeated measurements under the same loading condition (in this case,  repeated biaxial tests), or by introducing multiple loading modes, including a combination of biaxial and uniaxial tests in the $X$-direction (two tests), and biaxial and uniaxial tests in both the $X$- and $Y$-directions (three tests).}
    \label{noise_example} 
\end{figure}

\subsection{Performance in the presence of noise}

% Fig 4
In the previous synthetic examples, data was assumed to be noise-free.  However,  we anticipate noise shall have an effect on the final results. Hence, as the next synthetic example, we illustrate this fact by introducing noise, which can be interpreted as the error of measurement in experimental data. To this end, we consider the case where the noise is introduced in displacement measurement, i.e., ${{\bf{u}}^{{\rm{noise}}}} = {\bf{u}} + {\bf{\epsilon}}$ with ${\bf{\epsilon}}$ denoting Gaussian noise with mean and variance $(\mu,\sigma)$. In this example, we consistently assume $\mu=0$ and three cases of noise: noiseless, $\sigma=0.0007$, and $\sigma=0.001$. Moreover, we consider a square plate consisting of a $50\, \rm{by}\, 50$ quadrilateral mesh with a ring-type feature. Additionally, we consider five steps of uniform loading with $\lambda_{steps}=[1.05544, 1.07795, 1.10065, 1.12326, 1.14587]$.     Fig \ref{noise_example}a shows the generating material and corresponding displacement fields from the finite element simulation at $\lambda({\bf X})=1.14587$. Fig \ref{noise_example}b show the strain field interpolated by a neural network with Fourier features after injecting different levels of noise to the displacement field. The corresponding error with respect to ground truth data  increases by increasing $\sigma$, as expected, but with errors in the strain being an order of magnitude larger than those in the displacements.  Fig \ref{noise_example}d shows the result of the inverse problem. The inverse analysis is sensitive to the noise in the data. In agreement with previous examples, the inverse problem works well in the noiseless case, but the method failed to accurately recognize the non-homogeneity for both $\sigma=0.001,0.0007$. We remark that the noise of $\sigma=0.0007$ means that the 99 percent confidence interval of the displacement variation with respect to ground truth $[-0.001806,0.001806]$, while the noiseless displacements are on order $~0.15$. Note also that this error leads to $6$\% error in $\bf{F}$, as shown in Fig \ref{noise_example}c. Thus, about $0.06$ error is tolerated in the non-dimensional input $\bf{F}$. This is the metric that should be assessed to anticipate whether or not the inverse problem is likely to work accurately. Another metric that can be used in this regard is the signal-to-noise ratio (SNR), which can be defined as:
\begin{equation}\label{eq:SNR}
\begin{aligned}
{\rm{SN}}{{\rm{R}}_j} = 10\,{\log _{10}}\left( {\frac{{\sum\limits_i {{{\left\| {{{\bf{F}}_{{\rm{noiseless}}}}\left( {{{\bf{X}}_i},{t_j}} \right)} \right\|}^2}} }}{{\sum\limits_i {{{\left\| {{{\bf{F}}_{{\rm{noisy}}}}\left( {{{\bf{X}}_i},{t_j}} \right) - {{\bf{F}}_{{\rm{noiseless}}}}\left( {{{\bf{X}}_i},{t_j}} \right)} \right\|}^2}} }}} \right). 
\end{aligned}
\end{equation} 
The computed SNR values corresponding to the results in Fig \ref{noise_example}b are $19.5466$ and $18.9434$ for the cases  with $\sigma = 0.0007$ and $\sigma = 0.001$, respectively.    \\

In Figs \ref{noise_example}e, we aim to show that the inverse problem works even in the presence of noisy data if enough data is available. One strategy is increasing the number of noisy samples of the same biaxial test. Fig \ref{noise_example}-e shows the outcome of the inverse problem considering two or three samples in which the displacement has been corrupted with Gaussian noise with variance $\sigma=0.0007$. Clearly, a larger number of samples of the same experiment can be effectively used to filter out the noise. Note that the Gaussian noise is added to the displacements, but the data that goes into the model is the interpolated deformation gradient, which may not have Gaussian noise.  Alternatively, increasing the type of tests may improve the prediction as illustrated in Fig \ref{noise_example}e. These results show that if, in addition to an equibiaxial test, one also performed a uniaxial test on the same sample, the inverse identification of the inhomogeneity would improve. A set of three tests, equibiaxial and two uniaxial cases would further improve the inverse identification of the heterogeneous material. It is worth mentioning that we employed the same architecture used to generate results in Fig \ref{p_example}.

% We show the different magnitudes of $u$ that will, of course, break the inference. While we can still recover some result with noise of $xxx$, up to $xxx$ breaks the inference. Remark that the noise of $xxx$ means that displacements with confidence interval $[X,X]$ is added, and the noisless displacements are on the order $XXX$. Note also that this error leads to $xxxx$ error in F. On the other hand, the error of XX to displacement leads to error XXX in $F$. So about $XXX$ error tolerated in F which is nondimensional and much more significant than the error in u. 

% But how to deal with error is with good denoising or with more data. Not shown in the figure but in the Supplement the F interplation with NN can get rid of noise, and this is controlled by the neural network, in particular by the fourier features chosen which can smooth out higher frequencies depending on parameters. But what we do illustrate in the main Figure X is that circumvent with more data. For example, if the experiment is repeated multiple times and the noise is Gaussian of course averaging multiple experiments should denoise. Otherwise using all three samples as three different observations for the loss which is what is ullstrated. The reason why we opt for the multiple observations of the same experiment is because we want a fiar comparison against the case where we have different experiments. Thats illustrated in Fig X. adding uniaxial in two directions to the biaxial data. Both  multiple repeats as well as different test improve the performance. 
\begin{figure}[h!]
    \centering
    % \subfile{figures/fig_architecture}
        \begin{adjustbox}{center}
            \includegraphics[width=1.2\textwidth]{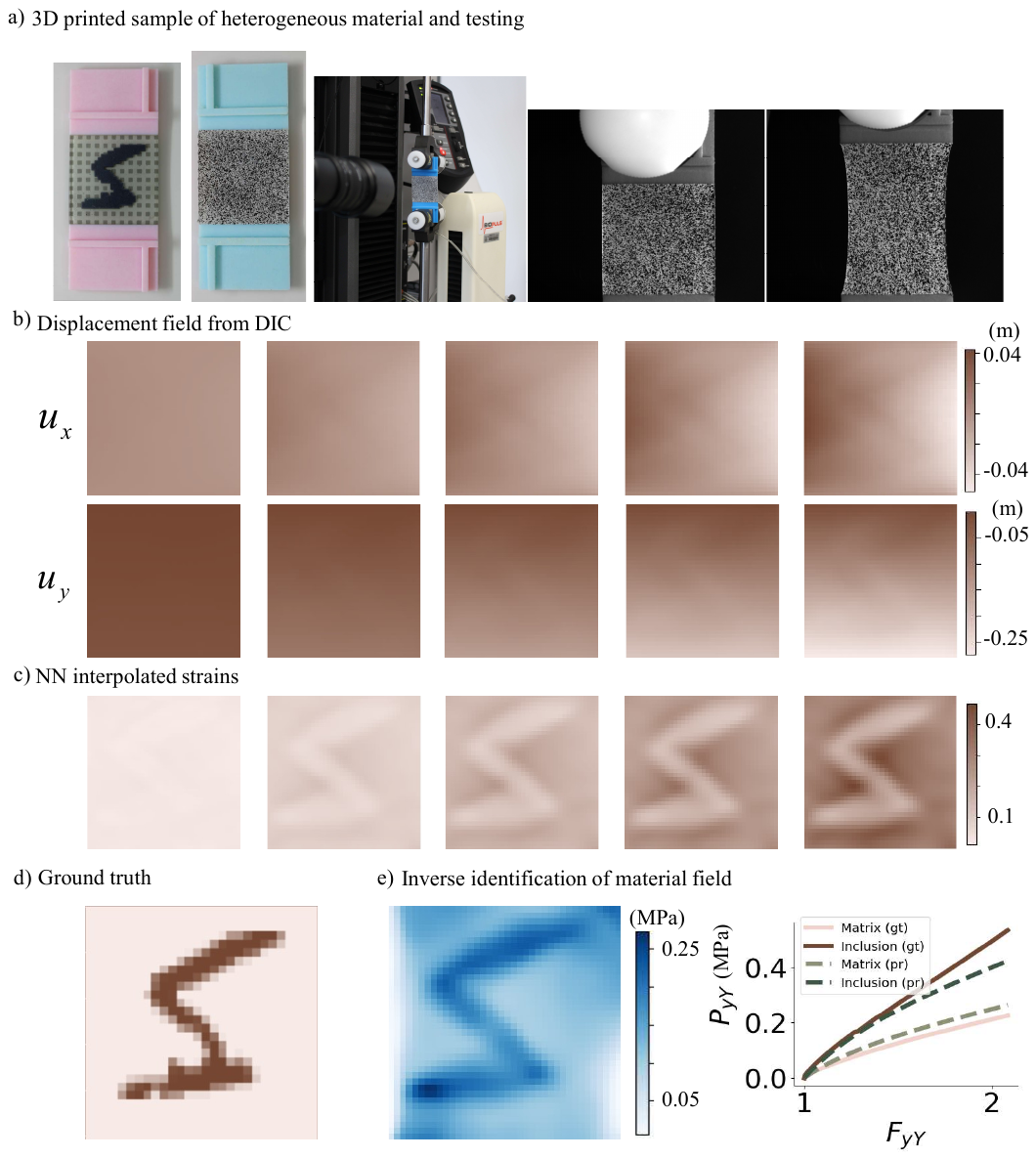}
    \end{adjustbox}
    \caption{Characterizing heterogeneous materials from experimental data on 3D printed elastomers.
(a) Experimental setup, where a square specimen contains an embedded inhomogeneity that is 3d printed using geometries from the MNIST dataset with a specified material. (b) Corresponding displacement fields obtained through  DIC, and (c)  the resulting maximum principal strain fields computed using the neural network-based interpolation.  (d) illustrates the performance of the proposed inverse method in identifying the material inhomogeneity within the domain and predicting the constitutive behavior of both the inclusion and the surrounding matrix.}
    \label{experimental_example} 
\end{figure}

\subsection{Demonstration with experimental data based on the MNIST geometries}

% Fig 5
In the last example we no longer consider synthetic examples but rather work with experimental data. With a Stratasys PolyJet 3D printer, we printed a heterogeneous geometry from the MNIST dataset. The printed geometry is illustrated in Fig \ref{experimental_example}-a. The inclusion is made out of a material with approximate tangent modulus $\mu=1400$ kPa, while the surrounding material has tangent modulus $\mu=550$ kPa.  The heterogeneous material was speckled, mounted on a tensile testing machine, and tested uniaxially as shown. DIC software LaVision DaVis was used to process the data and obtain displacement fields (Fig \ref{experimental_example}-b). The deformation gradient was computed on the discretized grid with the finite difference method and then interpolated with neural network as illustrated in Fig \ref{experimental_example}-c . Similar to the synthetic examples, the interpolated $\bf{F}(\bf{X})$ was used in the inverse model framework, together with the boundary conditions. The material field from the MNIST image and the inferred material field ($P_{yY}$ at $\lambda_x(\bf{X})=1.198$) are reported  in Fig \ref{experimental_example}-d \& -e, respectively. Furthermore, because we have a complete NODE material model per point, Fig \ref{experimental_example}-e shows the material behavior under uniaxial tension for two different points (one point to characterize matrix and the other one for the inclusion). The ground truth is the response from uniaxial tests of homogeneous samples tested in the same manner as the heterogeneous material. As can be seen, the method can discover the heterogeneity and the full material response in the case of experimental data. The architectures of the MLPs used to simulate the experimental results are as follows: [201, 40, 40, 4] with an initial Fourier feature layer of size [2, 100] for interpolating the deformation gradient field $\mathbf{F}$, which was numerically computed using the FDM from the displacement field $\mathbf{u}$ exported via DIC. Similar to previous examples, the MLPs used at the core of the NODEs have architecture [1, 4, 4, 4, 1]. Finally, the MLP hyper-network has architecture [200, 100, 50, 50, 50], with an initial Fourier feature layer of size [2, 100]. 

\section{Discussion}

We presented an inverse identification approach that reconstructs stress and heterogeneous material response fields from deformation gradient or displacement data. A key contribution of our method is that we output a data-driven material field, which is inferred without assuming a predefined constitutive model or prior knowledge of its spatial distribution. Instead, the material field is represented using a field of NODE-based constitutive models, implemented through a hyper-network framework. This approach effectively captures complex, heterogeneous material responses and has broad applicability, as demonstrated through both synthetic and experimental data.

One key takeaway from the synthetic and experimental examples is the method’s ability to identify a wide range of material heterogeneities, including complex geometries such as a $P$ shape, an $X$, an elliptical inclusion, a ring, and even an MNIST-derived heterogeneous material. This demonstrates that the approach imposes minimal restrictions on the types of heterogeneity it can capture. In terms of material response, the isotropic-anisotropic example in Fig  \ref{anisotropy_example} best highlights the method’s flexibility. Many methods for inverse parameter identification are based on strong assumptions, either regarding the geometry or the material behavior. Standard methods such as the virtual fields method works extremely well for homogeneous materials with a known material model form, but struggles to work robustly for arbitrary heterogeneous materials \cite{grediac2006virtual,avril2008overview,sutton2008identification}. Another, standard alternative, is finite element updating, which also imposes strong assumptions on the material model and spatial distribution of the material \cite{ereiz2022review,pierron2012virtual}. 

Novel machine-learning methods for material identification have been proposed in recent years, with similarities and differences with respect to the method proposed here. In \cite{valdes2022phase}, given a strain field, an optimization problem is setup to find an equilibrium stress field assuming the presence of a small number of phases. A material model is not explicitly sought, just stress-strain pairs for each phase. Similar to the data-driven approach \cite{valdes2022phase}, we also aim to retrieve an equilibrium stress field, see Fig \ref{p_example}. However, we do not enforce a discrete set of material responses. Additionally, we  predict a full material model at every point. Other deep learning methods such as in \cite{wu2025identifying,liu2022deep,chen2021learning,kirchhoff2024inference}, impose a closed form for the material, e.g. linear elastic or neo-Hookean, but make no assumptions regarding the spatial heterogeneity. In \cite{shi2025deep}, similar to \cite{valdes2022phase}, the deep learning framework first identifies different regions corresponding to different materials, reconstructs the equilibrium stress-field, and further uses data-driven ICNNs to describe the materials rather than enforcing a particular constitutive model. Similar to \cite{shi2025deep}, our constitutive model framework is data-driven, but based on NODEs rather than ICNNs \cite{tac2022node,padmanabha2024improving}. In contrast to \cite{shi2025deep}, our hyper-network approach results in a full NODE-field. 

An alternative approach to inverse problems based on artificial intelligence that has been developed recently is based on the interpolation of high dimensional probability distributions with generative methods, followed by  conditional sampling given a new strain field, e.g. \cite{patel2022solution}. In our case, direct minimization of the strong form yields only one solution in an inherently ill-posed inverse problem. Thus, for cases where information is known about the material field that can be captured with an appropriate prior distribution, then a Bayesian approach such as \cite{patel2022solution} would be better. For instance, in the last example, Fig \ref{experimental_example}, we could have used the mechanical MNIST dataset as a prior to regularize the geometry of the inverse problem \cite{lejeune2020mechanical}. For the material, restricting the prior is not necessarily confining ourselves to closed form material model. We have shown the ability to leverage generative models to describe probabilities over NODE materials \cite{tac2025generative}. 

We observed sensitivity to error or noise in the displacement and strain fields in Fig \ref{noise_example}. Errors in strain or deformation gradient of up to 6 perfect were handled well in the inverse identification. Beyond that level of noise we showed that it is possible to recover the material field if more data is added, either additional repeats of the same test, or additional testing methods. DIC methods are very well established and errors are low, tenth's of a pixel in displacement which commonly leads to less than five percent errors on strains \cite{siebert2007error,bornert2009assessment}.  Our experimental setup, Fig \ref{experimental_example}, shows that the method can work with experimental DIC data. Improvements could be done without additional data in ways already mentioned, e.g.  by restricting the number of materials, or by restricting the geometry or constitutive models with a strong prior.

\section{Conclusions}

In this work, we introduced an inverse identification approach for reconstructing stress and heterogeneous material response fields directly from deformation gradient or displacement field data. A key feature of our method is its data-driven nature, which allows for material inference without imposing predefined constitutive models or spatial distributions. By leveraging a hyper-network framework with NODE-based constitutive models, our approach effectively captures complex, heterogeneous material behaviors across a wide range of scenarios. This method is particularly relevant for material identification in tissues, composites, welded materials, among others, where full-field measurements should
be used to identify heterogeneous properties with complex geometric variation.

\section{Methods}\label{method_sec}

\subsection{Inverse problem setup}

Let denote $\mathbf{X}\in \mathcal{B}_0$ and  $\mathbf{x} = \varphi (\mathbf{X}) \in \mathcal{B}$ as undeformed and deformed configurations of a hyperelastic solid, respectively. The deformation gradient is defined as $\mathbf{F}=\nabla_{\mathbf{X}} \varphi$, and  $\mathbf{C}=\mathbf{F}^T\mathbf{F}$ and  $\mathbf{E}=(\mathbf{C}-\mathbf{I})/2$ are right Cauchy Green and Lagrange strain tensor, respectively. Steady-state balance laws of continuum mechanics in the reference configuration impose the following boundary value problem 

\begin{equation}\label{eq:equilibrium}
\begin{aligned}
&\nabla_\mathbf{X}\cdot \mathbf{P}=\mathbf{0}\quad \mathrm{in} \quad \mathcal{B}_0\\
& \mathbf{P} \mathbf{F}^T= \mathbf{F} \mathbf{P}^T \quad \mathrm{in} \quad \mathcal{B}_0
\end{aligned}
\end{equation} 
with boundary conditions 
\begin{align*}
    &\mathbf{x} = \bar{\mathbf{x}} \quad \mathrm{in} \quad \Gamma_u \\
    &\mathbf{P}\cdot \mathbf{N}=\bar{\mathbf{T}} \quad \mathrm{in} \quad \Gamma_t 
\end{align*}
where $\mathbf{P}$ is the first Piola Kirchhoff stress tensor, $\mathbf{N}$ is the vector normal to the surface in the reference configuration, $\Gamma_u$ is the portion of the boundary where Dirichlet boundary conditions are applied and $\Gamma_t$ is the portion of the boundary where traction forces are applied. 

%Note that the above boundary value problem can also be formulated in the deformed domain $\mathcal{B}$ in terms of the Cauchy stress tensor $\boldsymbol{\sigma}$.  
To close the system of equations, a constitutive equation is needed to relate the deformation gradient $\mathbf{F}$ (or the strains) to the stress $\mathbf{P}$. For hyperelastic materials this relationship depends on a single scalar potential, the strain energy density function, by

\begin{equation}\label{constituive_eq}
    \mathbf{P} = \frac{\partial \Psi(\mathbf{F},\bf{X})}{\partial \mathbf{F}},
\end{equation}
where we assumed the strain energy is spatial dependence to account for inhomogeneity. 
Now, assume that we have an strain field $\bf{E}(\bf{X})$ ---either directly from DIC or after approximating it from the displacement field using a numerical method such as FDM--- and forcing boundary conditions $\mathbf{\bar T}$ on the boundary $\Gamma_t$. The goal of the data-driven inverse problem is to learn strain energy density $\Psi(\mathbf{F},\bf{X})$ that correctly relates the deformation gradient to stress $\mathbf{P}(\mathbf{F},\bf{X})$ that satisfies the boundary value problem \eqref{eq:equilibrium}.
%assuming a hyperelastic material, then we want to learn a strain energy field $\Psi(\mathbf{E},\mathbf{X})$. 

In classical inverse methods, the above problem is usually treated by selecting a family of constitutive equations, e.g., the neo-Hookean model, and then the problem is re-cast into a minimization problem whose parameters are the constants of that specific model. The minimization problem in these methods is usually based on a variational form of \eqref{eq:equilibrium}, and thus, the satisfaction of the equilibrium equations is through a weak sense. The associated integral over the domain in these approaches is carried out by approximating the fields with a set of compactly supported function spaces similar to FEM. \\
However, selecting a specific class of constitutive equations imposes restrictions on material identification in classical methods. In fact, every analytical constitutive model has been designed to describe the behavior of a particular class of materials under specific conditions. As a result, the accuracy of material identification heavily depends on the expertise of the specialist regarding the constitutive equations. Furthermore, the minimization problem in these methods involves finding the minimizer of a functional, which is implemented through discretization and a corresponding set of compactly supported functions. Consequently, the solution inherently depends on the mesh size and the type of shape functions employed therein.

In previous work, we introduced a data-driven model with NODEs with alternatives, such as ICNN or CANNs \cite{tacDatadrivenAnisotropicFinite2023}. These models make no assumption on the format of strain energy and are extremely flexible and accurate. They are overparameterized by a large number of weights and biases $\boldsymbol{\theta}$. The inverse problem then is to estimate field of neural network parameters $\boldsymbol{\theta}(\bf{X})$.  This approach would allow solving the inverse problem when there is potentially different material behaviors in different parts of the domain.

Before ending this section, we briefly outline conditions on the format of the strain density functions, which are dictated by laws of physics and mathematics. Firstly, any constitutive equations should be frame-indifferent, which imposes the following format on the strain density energy $\Psi ({\bf{F}},{\bf{X}}) = \hat \Psi ({\bf{C}},{\bf{X}})$. Also, the strain energy should always be positive for any admissible deformation and satisfy the vanishing condition in the natural state, that is, $ \Psi ({\bf{F}},{\bf{X}}) \ge 0,\, \Psi ({\bf{I}},{\bf{X}}) = 0,\,{\left. {\frac{{\partial \Psi }}{{\partial {\bf{F}}}}} \right|_{{\bf{F}} = {\bf{I}}}} = {\bf{0}}$, where $\bf{I}$ is the identity tensor. Also, to explain the concept of material symmetry, we need to define what an isotropic function is:
\begin{definition}
Given a list of second-order tensors ${\mathbf{A}}_1, ..., {\mathbf{A}}_n$, a scalar function $\Psi ({{\bf{A}}_1},...,{{\bf{A}}_n},{\bf{X}})$ is called isotropic function if:
\begin{equation}
    \Psi^{iso} ({\bf{Q}}{{\bf{A}}_1}{{\bf{Q}}^T},...,{\bf{Q}}{{\bf{A}}_n}{{\bf{Q}}^T},{\bf{X}}) = \Psi^{iso} ({{\bf{A}}_1},...,{{\bf{A}}_n},{\bf{X}}),\,\,\forall {\bf{Q}} \in {\bf{O}}{\rm{rth}}
\end{equation}
\end{definition}
Now, when the material's behavior is invariant under any rotation, the strain energy density is ${{\hat \Psi }^{{\rm{iso}}}}({\bf{C}},{\bf{X}})$, which can be shown that it degenerates to a function of invariants of the right Cauchy-Green strain tensor, i.e.,  $\Psi ({\bf{F}},{\bf{X}}) = \hat \Psi ({{\bf{I}}_{\bf{C}}},{\bf{I}}{{\bf{I}}_{\bf{C}}},{\bf{II}}{{\bf{I}}_{\bf{C}}},{\bf{X}})$.
\begin{definition}
Let a symmetry group be represented by a set of second-order tensors ${\bf{L}}_i$, $i = 1, \dots, n$, known as structural tensors (see~\cite{itskov2007tensor}), that is:
\begin{equation}
\mathcal{G}^s = \left\{ {{\bf{Q}} \in {\bf{O}}{\rm{rth}}:\,{\bf{Q}}{{\bf{L}}_i}{{\bf{Q}}^T} = {{\bf{L}}_i},i=1,...,n} \right\}, 
\end{equation}
Then, a material is said to be anisotropic with group symmetry $\mathcal{G}^s$ if the following holds:\\
 \begin{equation}
\hat \Psi^{aniso} ({\bf{QC}}{{\bf{Q}}^T},{\bf{X}}) = \hat \Psi^{aniso} ({\bf{C}},{\bf{X}}),\,\,\,{\bf{Q}} \in \mathcal{G}^s 
\end{equation}
\end{definition}

 It can be shown that any anisotropic strain density function defined above is characterized by an isotropic function of the right Cauchy-Green strain and the corresponding structural tensors (\cite{zhang1990structural, itskov2007tensor}): ${{\hat \Psi }^{{\rm{aniso}}}}({\bf{C}},{\bf{X}}) = {{\hat \Psi }^{{\rm{iso}}}}({\bf{C}},{{\bf{L}}_1},...,{{\bf{L}}_n},{\bf{X}})$. From this point onward, for the sake of brevity, we denote isotropic functions with $\Psi$. The next concept, which has a mathematical background rather than a physical one, is the notion of polyconvexity \cite{ball1976convexity}. Specifically, this property ensures that the strain energy satisfies the ellipticity of the acoustic tensor \cite{hartmann2003polyconvexity}, and thus, the displacement wave speed is always a real number--- which, together with the growth condition,  ensures the existence of a global minimizer. Mathematically, a strain  density energy is polyconvex if and only if
$\Psi ({\bf{F}},{\bf{X}}) = {\Psi ^{\mathcal{P}}}({\bf{F}},\,{\rm{cof}}\,{\bf{F}},\,\det \,{\bf{F}},{\bf{X}})$, where $\Psi ^{\mathcal{P}}$ is convex with respect to first three arguments \cite{ball1976convexity, schroder2003invariant}. Finally, in the language of data-driven methods, a constitutive is called thermodynamically consistent if  \eqref{constituive_eq} holds. Also, it is worth noting that  $\eqref{eq:equilibrium}_2$ is satisfied if we consider the following:
\begin{equation}\label{const-symmetry}
    {\bf{S}} = 2\frac{{\partial \hat \Psi ({\bf{C}},{\bf{X}})}}{{\partial {\bf{C}}}}
\end{equation}
where ${\bf{S}}={\bf{S}}^T$ is the second Piola-kirchhoff stress, which is related to the first Piola-kirchhoff by ${\bf{P}} = {\bf{FS}}$.    
% quick paragraph polyconvex NODE
% \subsection{Neural ordinary differential equations}

% NODEs are machine learning frameworks such that the output is defined as the solution of an ordinary differential equation (ODE) or a system of ODEs at a given time. 
% \begin{align*}
%     % \label{eq_ODE}
%     \frac{d\mathbf{h}(\tau)}{d\tau} = \mathbf{f}_{\text{NODE}}(\mathbf{h}(\tau), \tau, \boldsymbol{\varphi}) \, ,
% \end{align*}
% where $\tau$ is a pseudo time variable, and $\mathbf{h}$ the variables of interest. The right hand side of the ODE, $\mathbf{f}_{\text{NODE}}(\cdot, \cdot, \boldsymbol{\varphi})$, is a MLP parameterized by $\boldsymbol{\varphi}$. From the fundamentals of ODEs we know that the solution trajectories of ODEs do not intersect, provided that the right hand side is Lipschitz continuous. For the scalar variable case, $h(\tau)$, this implies that for trajectories $h_1$ and $h_2$, the following holds
% \begin{align*}
%     h_1(0) \geq h_2(0) \Longleftrightarrow h_1(1) \geq h_2(1)\, ,
%     % \\
%     % h_1(0) \leq h_2(0) \Longleftrightarrow h_1(1) \leq h_2(1)
% \end{align*}
% which means that the input-output map of a scalar NODE is monotonic. We employ this property of NODEs to construct polyconvex strain energy density functions as we explain in the next subsection.
\subsection{Deep learning architectures}
Now that the main goal of the inverse problem has been elaborated and the conditions on constitutive equations have been outlined, we explain the key steps of the proposed method in this section. \\
In principle, a NODE is a generalization of the residual network \cite{chen2018neural}. More specifically, to approximate a target function $g(X)$,  a first-order ordinary differential equations (ODEs) is solved
\begin{equation}\label{NODE_def}
\begin{aligned}
&\frac{{dh}}{{dt}} = f(h,t;\theta ),\  0 \le t \le 1\\
&h(0) = X, h(1)=g(X)
\end{aligned}
\end{equation}
where $f$ is a function with parameters denoted by $\theta$. Picard–Lindelöf theorem implies the uniqueness of the solution of the above ODE when  $f$ is a Lipchitz continuous function with respect to $h$ and continuous in $t$, which is the case when $f$ is substituted by a multilayer perceptron network (MLP). This fact implies that the trajectory governed by the above ODE is nonintersecting, and the resulting function is invertible and thus monotonic. As a result, if we select $g(0)=0$---which can be achieved with a minor modification to the neural network architecture \cite{tac2022node, tacDatadrivenAnisotropicFinite2023}--then monotonicity implies: $X > 0,\,\,g(X) > 0$, leading to the convexity of  the function $G(X)$ defined by $G(X) = \int\limits_0^X {g(y)dy} $. This property can be leveraged to build neural networks that preserves polyconvexity property of strain density energy functions. In particular, by using invariants that  inherently satisfy polyconvexity condition as the input of the above-defined NODEs, we can make sure that the following is a generic form to construct polyconvex functions:
\begin{equation}\label{NODE_potential}
\begin{aligned}
&\frac{{dh}}{{dt}} = f(h,t;\theta ),\  0 \le t \le 1\\
&h(0) = {\bf{X}},\,\,\,\,h(0) = \frac{d\psi }{dI_{\rm{poly}}}
\end{aligned}
\end{equation}
where $I_{\rm{poly}}$ stands for an invariant fulfilling polyconvexity (see \cite{tac2022node} for more details). Hence,  we utilize strain energy density functions of the following form, in which every right hand side function is defined based on  \eqref{NODE_potential} with $f$ being replaced by a separate MLPs, \cite{tac2022node}: 

\begin{equation}    \label{eq_strain energy density function_general}
\begin{aligned}
{\Psi ^{{\rm{NODE}}}} =& {\Psi _{{I_1}}}({I_1};\theta _{{I_1}}^{{\rm{NODE}}}({\bf{X}})) + {\Psi _{{I_2}}}({I_2};\theta _{{I_2}}^{{\rm{NODE}}}({\bf{X}}))\\& 
+ {\Psi _{{I_{4v}}}}({I_{4v}};\theta _{{I_{4v}}}^{{\rm{NODE}}}({\bf{X}}))+ {\Psi _{{I_{4w}}}}({I_{4w}};\theta _{{I_{4w}}}^{{\rm{NODE}}}({\bf{X}})) \\
&  + \sum\limits_{j > i} {{\Psi _{{I_i},{I_j}}}} \left( {{\alpha _{ij}}{I_i} + (1 - {\alpha _{ij}}){I_j};\theta _{{\alpha _{ij}}}^{{\rm{NODE}}}({\bf{X}})} \right) + {\Psi _J}(J;\theta _J^{{\rm{NODE}}}({\bf{X}})),\\
&i= I_1,I_2,I_{4v}, I_{4w}, j=I_2,I_{4v}, I_{4w}
\end{aligned}  
\end{equation}
where $I_1, I_2, I_{4v}$ and $I_{4w}$ are polyconvex invariants of the right Cauchy-Green deformation tensor, $\mathbf{C}$, and $J=\det \mathbf{F}$. It is noted that to account for heterogeneity, we assume the parameters of the NODEs for right hand side functions are spatially dependent, which are defined by a hyper-network whose input is material coordinates $\mathbf{X}$ and the output is the parameters of all NODEs (see Fig \ref{fig_summary}):
\begin{equation}    \label{hyper-network}
\begin{aligned}
{\rm{vec}}\left( {\theta _{{I_1}}^{{\rm{NODE}}},\,\theta _{{I_2}}^{{\rm{NODE}}},\,\theta _{{I_{4v}}}^{{\rm{NODE}}},\,\theta _{{I_{4w}}}^{{\rm{NODE}}},\,\theta _{{\alpha _{ij}}}^{{\rm{NODE}}},\,\theta _J^{{\rm{NODE}}}} \right) = N{N^{{\rm{hyper}}}}({\bf{X}};{\theta ^{{\rm{hyper}}}}), 
\end{aligned}  
\end{equation}
in which $\rm{vec}(\theta)$ returns the vectorized format of any list of parameters $\theta$. We can define predicted first Piola–Kirchhoff as:
\begin{equation}\label{Node_constitutive_predicting}
\begin{aligned}
{{\bf{P}}^{{\rm{pred}}}} = 2{{\bf{F}}^{{\mathop{\rm int}} }}\frac{{\partial {\Psi ^{{\rm{NODE}}}}}}{{\partial {{\bf{C}}^{{\mathop{\rm int}} }}}} = 2{{\bf{F}}^{{\mathop{\rm int}} }}\sum\limits_i^{} {\frac{{\partial {\Psi ^{{\rm{NODE}}}}}}{{\partial I_i^{{\mathop{\rm int}} }}}} \frac{{\partial I_i^{{\mathop{\rm int}} }}}{{\partial {{\bf{C}}^{{\mathop{\rm int}} }}}}, i = 1,2,\,\,4v,\,\,4w,\,\,{{\bf{C}}^{{\mathop{\rm int}} }} = {\left( {{{\bf{F}}^{{\mathop{\rm int}} }}} \right)^T}{{\bf{F}}^{{\mathop{\rm int}} }} 
\end{aligned}  
\end{equation}
where ${{\bf{F}}^{\rm{int}}}$ is deformation gradient field interpolated by a neural network, which shall be explained in the sequel. 
% We have shown that in order to preserve polyconvexity of the strain energy density function each of the $\Psi_{I_i}$ and $\Psi_{I_i, I_j}$ needs to be convex non-decreasing. This is equivalent to monotonicity of the derivatives, $d\Psi_{I_i}/d I_i$ with $d\Psi_{I_i}/d I_i \geq 0$ in the domain of $I_i$. As a result, modeling the derivative functions $d\Psi_{I_i}/d I_i$ with NODEs guarantees polyconvexity of the strain energy density function. 
% The NODE is 

% The hypernetwork is 

\subsection{Solution of the inverse problem}
The general steps of the method to solve the inverse problem have been illustrated in Fig \ref{fig_summary}. Assuming that a full field displacement obtained from the DIC setup is given as the input data, we apply a finite difference method, e.g., the central difference method,  to calculate the deformation gradient tensor. Alternatively, one can compute $\mathbf{F}$ with finite element basis functions. As the next step, an MLP is defined to interpolate the deformation gradient field whose input data is material coordinates and the output is the components of the deformation gradients, which is parametrized from the following optimization:
\begin{equation}    \label{eq_training_FDM_Fint}
\begin{aligned}
\mathop {\arg \min }\limits_{{\theta ^{{\rm{int}}}}} \frac{{\sum\limits_{{t_j} = 0}^m {\sum\limits_{i = 1}^{{n_p}} {{{\left\| {{{\bf{F}}^{{\rm{FDM}}}}({{\bf{X}}_i},{t_j}) - {{\bf{F}}^{{\rm{int}}}}({{\bf{X}}_i},{t_j};{\theta ^{{\rm{int}}}})} \right\|}^2}} } }}{{m{n_p}}}   
\end{aligned}  
\end{equation}
with $n_p$ and $m$ representing the number of spatial points and temporal instances, respectively,  and  ${{{\bf{F}}^{\rm{FDM}}}({{\bf{X}}_i},t_j)}$ and ${{{\bf{F}}^{\rm{int}}}({{\bf{X}}_i},t_j)}$ denote deformation gradient obtained from finite difference method at different time instances and the interpolated one,  respectively. At this stage, to initialize the parameters of the hyper-network and NODEs, we apply two pre-training procedures. First, we consider a homogenization scheme. In this regard, we define the spatial average of a tensorial quantity as:
\begin{equation}\label{pre-training0}
\begin{aligned}
\left\langle {\bf{A}} \right\rangle \left( t \right) = \frac{1}{{\left| \Omega  \right|}}\int\limits_\Omega  {{\bf{A}}\left( {{\bf{X}},t} \right)\,d\Omega } .  
\end{aligned}  
\end{equation}
Moreover, from (bi)axial experimental or synthetic data, one can write:
\begin{equation}    \label{pre-training1}
\begin{aligned}
P_{xX}^{{\rm{ave}}}\left( t \right) = \frac{{{f_x}\left( t \right)}}{{{L_y}}},P_{yY}^{{\rm{ave}}} = \frac{{{f_y}\left( t \right)}}{{{L_x}}},P_{yX}^{{\rm{ave}}} = P_{xY}^{{\rm{ave}}} = 0\,  
\end{aligned}  
\end{equation}
in which $f_x$ and $f_y$ are replaced with forces obtained either from DIC experiments for real data or resultant force from FEM analysis for synthetic data.\\
Now, by assuming that the parameters of NODEs are spatially independent, we initialize the parameters of NODEs with those obtained from the following homogenized minimization:
\begin{equation}    \label{pre-training1-3}
\begin{aligned}
\mathop {\arg \min }\limits_{{{\bar \theta }^{{\rm{NODE}}}}} \frac{{\sum\limits_{{t_j} = 0}^m {{{\left\| {{{\bf{P}}^{{\rm{ave}}}}\left( {{t_j}} \right) - 2\left\langle {{{\bf{F}}^{{\rm{int}}}}} \right\rangle \left( {{t_j}} \right)\frac{{\partial {\Psi ^{{\rm{NODE}}}}}}{{\partial \left\langle {{{\bf{C}}^{{\rm{int}}}}} \right\rangle \left( {{t_j}} \right)}}} \right\|}^2}} }}{m n_p} ,\left\langle {{{\bf{C}}^{{\mathop{\rm int}} }}} \right\rangle \left( {{t_j}} \right) = {\left\langle {{{\bf{F}}^{{\mathop{\rm int}} }}} \right\rangle ^T}\left( {{t_j}} \right)\left\langle {{{\bf{F}}^{{\mathop{\rm int}} }}} \right\rangle \left( {{t_j}} \right),   
\end{aligned}  
\end{equation}
where ${\bar \theta }^{\rm{NODE}}$ stands for average parameters of NODEs. Subsequently, a regular learning is performed to train the hyper-network to approximate this average field of variables for NODE's parameters.\\   
After the first initialization of NODEs' and the hypernetwork's parameters,  another pre-training procedure is implemented, specifically to initialize the parameters of the hyper-network. In doing so, We begin by introducing inhomogeneity into the constitutive equations such that, for a given non-uniform deformation gradient field, the corresponding stress field is ideally uniform and thus identically satisfies the equilibrium equations. While this defines a complex inverse problem that may not be satisfied point-wise, it serves as a good pretraining step for the main training, as regions with larger strains tend to correspond to softer constitutive responses, and regions with smaller strains to stiffer ones.     
% \begin{equation}    \label{pre-training1-4}
% \begin{aligned}
% {{\bf{P}}^{{\rm{ave}}}}\left( t \right) = 2\left\langle {{\bf{F}}\frac{{\partial \Psi }}{{\partial {\bf{C}}}}} \right\rangle \left( t \right).  
% \end{aligned}  
% \end{equation}
% with
% \begin{equation}    \label{pre-training1-5}
% \begin{aligned}
% \left\langle {{\bf{F}}\frac{{\partial \Psi }}{{\partial {\bf{C}}}}} \right\rangle  = \frac{1}{{\left| \Omega  \right|}}\int\limits_\Omega  {{\bf{F}}\frac{{\partial \Psi }}{{\partial {\bf{C}}}}d\Omega } .  
% \end{aligned}  
% \end{equation}
Hence, the second pre-training is to find parameters of the hyper-network such that the following is fulfilled:
\begin{equation}    \label{pre-training1-6}
\begin{aligned}
\mathop {\arg \min }\limits_{{\theta ^{{\rm{hyper}}}}} \,\,\,\,\,\frac{{\sum\limits_{{t_j} = 0}^m {\sum\limits_{i = 1}^{{n_p}} {{{\left\| {{{\bf{P}}^{{\rm{ave}}}}\left( {{t_j}} \right) - {{\bf{P}}^{{\rm{pred}}}}\left( {{{\bf{X}}_i},{t_j}} \right)} \right\|}^2}} } }}{{m{n_p}}}
\end{aligned}  
\end{equation}

% in the parameters of the NODEs, we assume that, which we tune   since  First 

Now, after two pre-training schemes,  we perform the main training, a multiobjective optimization consisting of equilibrium terms and forcing terms due to Neumann boundary conditions applied during experiments. In particular,  the parameters of the hyper-network, $\theta^{\rm{hyper}}$, are determined from the minimization of the following functional:
\begin{equation}\label{main_loss_function}
\begin{aligned}
\mathop {\arg \min }\limits_{{\theta ^{{\rm{hyper}}}}} \,\,\,\,\sum\limits_{{t_j} = 0}^m {\left( {\frac{{\sum\limits_{i = 1}^{{n_p}} {{{\left\| {{\nabla _{\bf{X}}} \cdot {{\bf{P}}^{{\rm{pre}}}}\left( {{{\bf{X}}_i},{t_j}} \right)} \right\|}^2}} }}{{m{n_p}}} + \lambda \sum\limits_{i = 1}^{{n_\Gamma }} {\frac{{{{\left\| {{{\bf{P}}^{{\rm{pre}}}}\left( {{\bf{X}}_i^\Gamma ,{t_j}} \right) \cdot {\bf{N}}\left( {{\bf{X}}_i^\Gamma } \right) - {\bf{t}}\left( {{\bf{X}}_i^\Gamma ,{t_j}} \right)} \right\|}^2}}}{{m{n_{{n_\Gamma }}}}}} } \right)}   ,  
\end{aligned}  
\end{equation} 
where $n_{\Gamma}$ is the total number of nodal boundary points on $\Gamma_t$,  $\bf t$ denotes prescribed traction boundary forces defined on $\Gamma_t$, and   $\lambda$ is a hyper-parameter. It is observed in \eqref{main_loss_function} that, similar to the classical inverse approach, a first-order differentiation is required---as the deformation gradient field is obtained through the above-mentioned interpolation scheme---while at the same time, the strong form of the equilibrium equations is incorporated into the loss function.    
\subsection{Synthetic data generation}
To generate synthetic data, we used JAX-FEM \cite{jaxFEM}. We considered the linear shape functions, and the JAX-FEM built-in function was used to calculate deformation gradients. Also, the geometry is a mesh of $50\,\, \rm{by}\,\, 50$ quadrilateral elements for a square with a dimension of unity.   

\subsection{Experimental data generation}
To generate experimental data, a heterogeneous sample was 3D-printed using a Stratasys PolyJet J750 Digital Anatomy printer (Stratasys, Eden Prairie, MN, USA). The geometry of the sample was taken directly from the MNIST dataset, and sample dimensions were $40 \times 40 \times 2$ mm. The sample consisted of two components: an inclusion and the surrounding material. The inclusion was composed of FLX9870 with a tangent modulus of about 1400 kPa, while the surrounding material was composed of Agilus30 with a tangent modulus of about 550 kPa. Additionally, to ensure a secure attachment between the sample and the mechanical testing device, rigid clamps were incorporated into the sample design. 

Following printing, the sample was prepared for DIC by applying a speckle pattern to one side. A thin base coat of white acrylic ink was first applied, followed by a distribution of small black dots to create a high-contrast pattern. The sample was then mounted onto a uniaxial tensile tester (Instron, Norwood, MA, USA) using pneumatic grips and extended until failure at a rate of 0.2 mm/s. A 100 N load cell (accuracy of $\pm$ 0.2 N) was used to measure force, while displacement and time data were recorded via Instron software. Furthermore, a custom LabVIEW program (National Instruments, Austin, TX, USA, Version 2021 SP1) was used to take images of the sample during extension at a rate of 5 Hz. These images were exported to LaVision DaVis (LaVision, G{\"o}ttingen, Germany, Version 11) for post-processing. Using the strain tracking feature, displacement fields were captured for a $42 \times 41$ grid covering the entire sample area.

\section{Acknowledgements}
% VT and IB acknowledge the support of AFOSR under the grant number FA09950-22-1-0061.
% FSC and MR acknowledge the support of the Open Seed Fund of the School of Engineering at Pontificia Universidad Cat\'olica de Chile. ABT acknowledges support from National Institute of Arthritis and Musculoskeletal and Skin Diseases, National Institute of Health, United States under award R01AR074525. 
ABT acknowledges support of ARO under award W911NF-24-1-0244. 

\section{Declarations}
The authors have no conflicts of interest to declare. 

\section{Supplementary material}
All data, model parameters and code associated with this study are available in a public Github repository at \url{https://github.com/tajtac/node_diffusion}.
\section{Appendix}
\subsection{Ablation study}
\begin{figure}[h!]
    \centering
    % \subfile{figures/fig_architecture}
        \begin{adjustbox}{center}
            \includegraphics[width=1.2\textwidth]{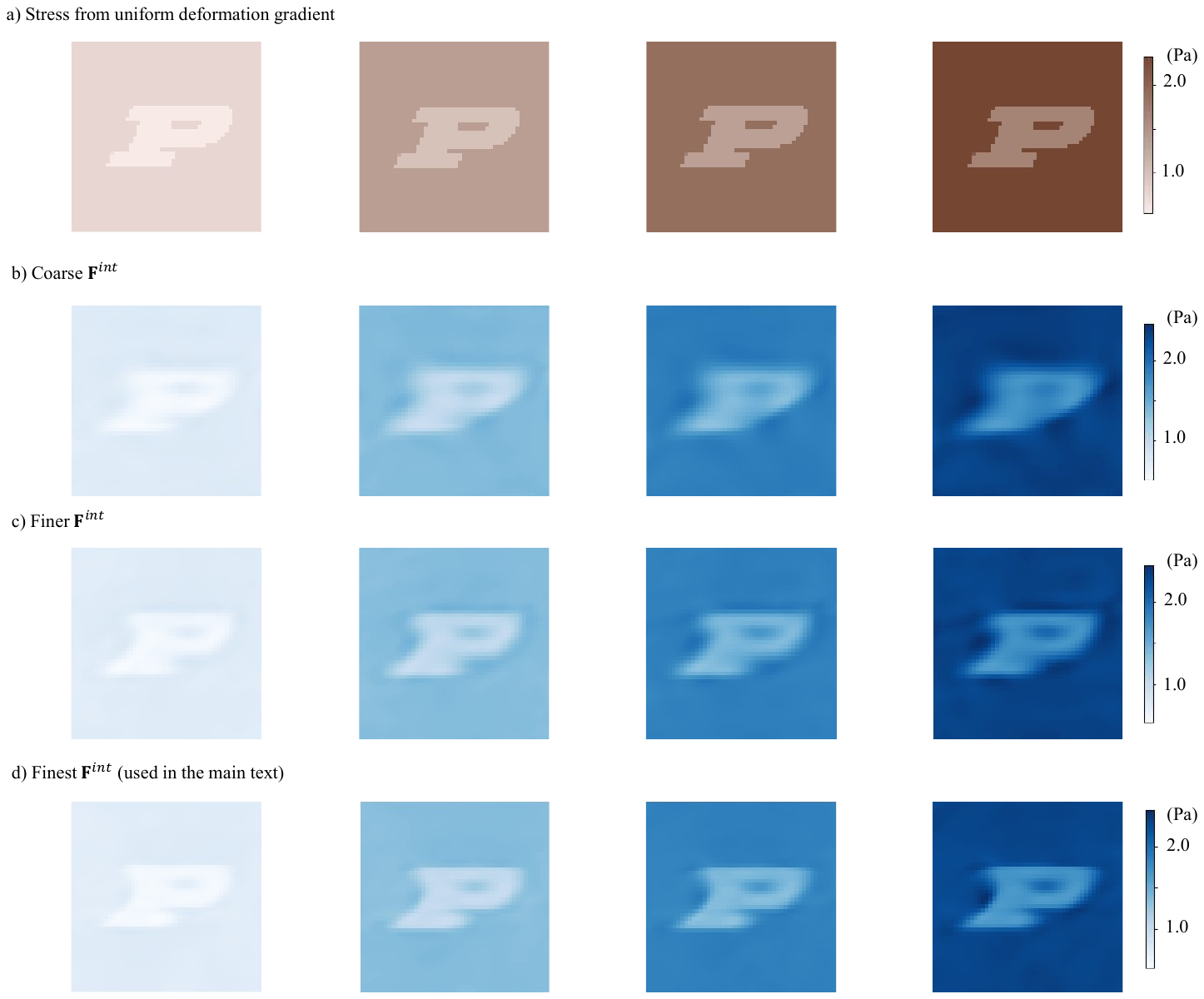}
    \end{adjustbox}
    \caption{Ablation study to demonstrate the effect of Fourier features used in the interpolation of ground truth data on the performance of the new inverse method. The ground truth stresses under constant strain fields are shown in (a). The corresponding results predicted by the method, using Fourier features with architectures [2, 10], [2, 40], and [2, 100] for interpolation, are shown in (b), (c), and (d), respectively.}
    \label{ablation_F_int} 
\end{figure}

\begin{figure}[h!]
    \centering
    % \subfile{figures/fig_architecture}
        \begin{adjustbox}{center}
            \includegraphics[width=1.2\textwidth]{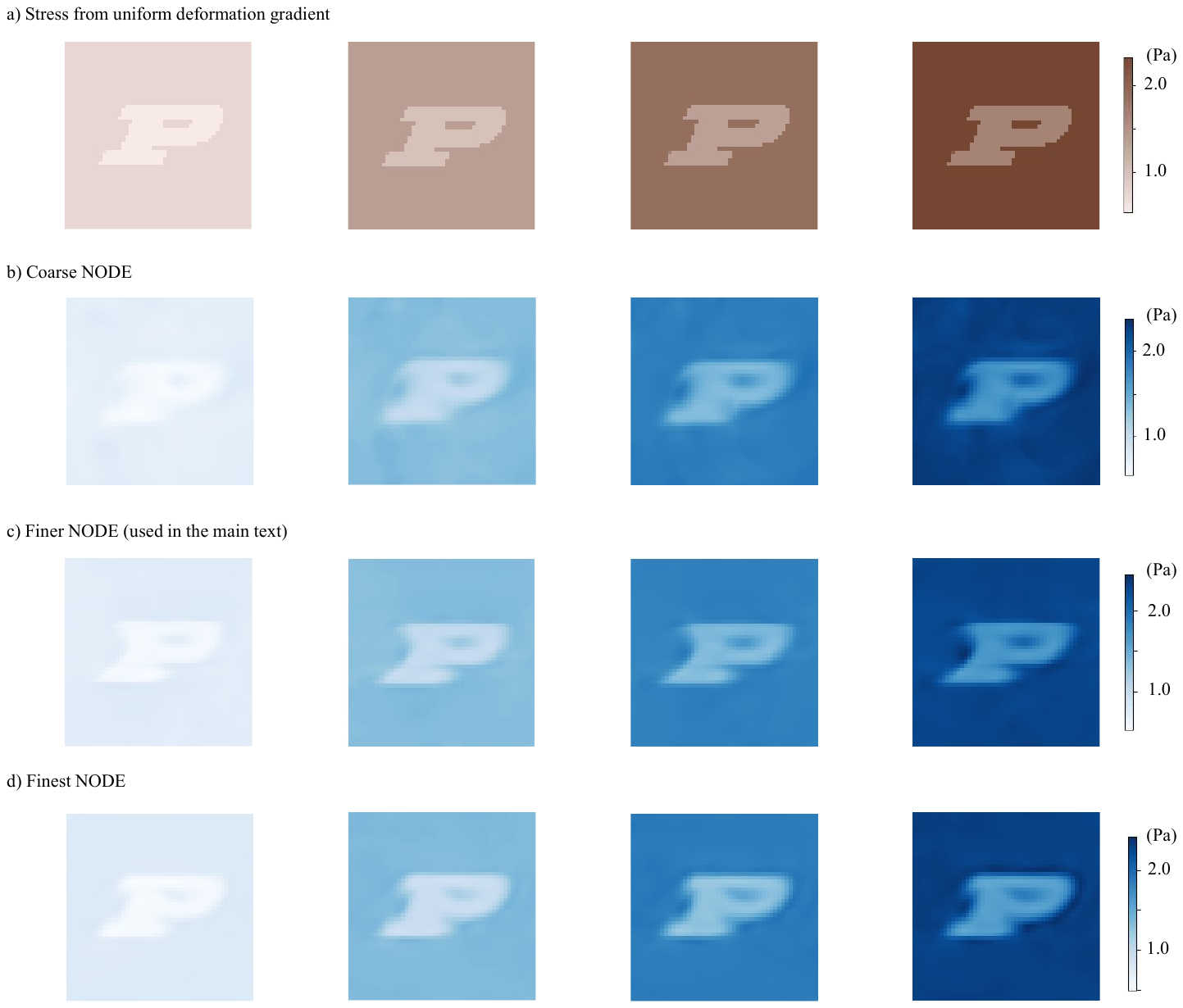}
    \end{adjustbox}
    \caption{Ablation study to show the effect of the architecture used in the core NODEs. The ground truth stresses under constant strain fields are shown in (a). Similar results corresponding to the architectures [1, 2, 1], [1, 4, 4, 4, 1], and [1, 8, 8, 8, 1] are shown in (b), (c), and (d), respectively.}
    \label{ablation_NODE} 
\end{figure}

\begin{figure}[h!]
    \centering
    % \subfile{figures/fig_architecture}
        \begin{adjustbox}{center}
            \includegraphics[width=1.2\textwidth]{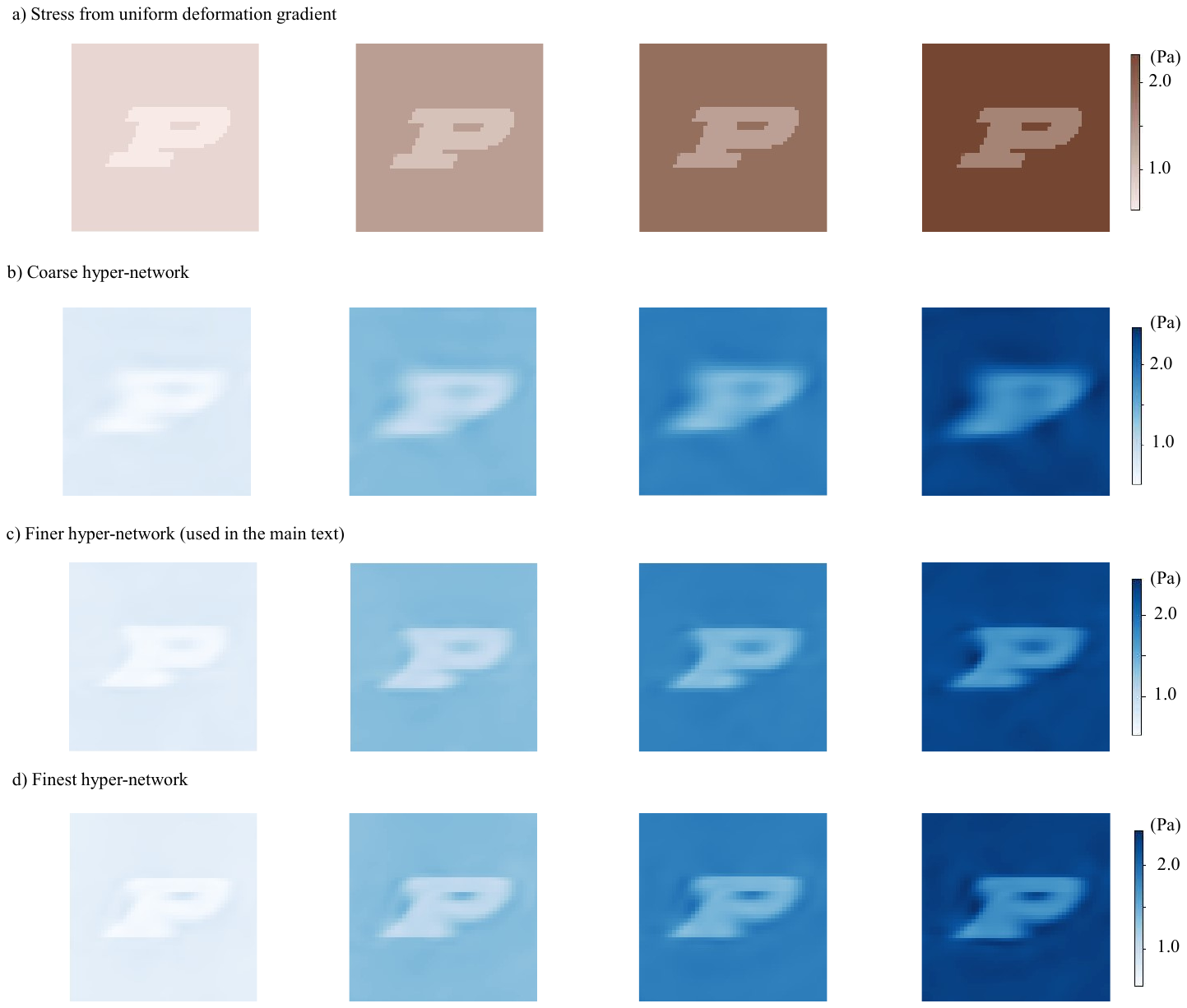}
    \end{adjustbox}
    \caption{Ablation study to indicate how various architecture used in hyper-network can affect the final outcome of the inverse problem. The ground truth stresses under constant strain fields are shown in (a). Panels (b), (c), and (d) show the corresponding results obtained using the architectures [20, 10, 126] with Fourier features [2, 10], [80, 40, 40, 126] with Fourier features [2, 40], and [80, 60, 60, 126] with Fourier features [2, 40], respectively.}
    \label{ablation_hyper} 
\end{figure}

In this section, we present an ablation study of the proposed method. In this regard, we consider the example with an embedded P-shape inhomogeneity. As discussed in Section\ref{method_sec}, the main components of the method include interpolation of the ground truth deformation gradient, a spatially dependent central NODE, and a hyper-network that defines the spatial variation of the NODE parameters across the domain. Hence, the ablation study we consider in this work is to investigate the effect of the above-mentioned components. For clarity, the ground truth data is shown in the first panel of each figure. To emphasize the differences, we consistently assess the accuracy of each configuration in identifying the inhomogeneity profile embedded in the square plate. To this end, we examine the nominal stress component ${{\bf{P}}}_{xX}$ evaluated under four levels of imposed strain: $\lambda = [1.05, 1.10, 1.15, 1.20]$. Moreover, the central architectures used in this ablation study are as follows: for interpolation of the deformation gradient, the central architecture is [201, 40, 40, 4] with Fourier features [2, 100]; for the NODEs, the central architecture is [1, 4, 4, 4, 1]; and for the hyper-network, the central architecture is [80, 40, 40, 126] with Fourier features [2, 40].\\
To obtain ${\bf{F}}^{\rm{int}}$, the key factor is incorporating Fourier features, which are essential for capturing sharp variations in the field. Accordingly, we present results for three architectures: [2,10]  (Fig \ref{ablation_F_int}a), [2,40] (Fig \ref{ablation_F_int}b), and [2,100] (Fig \ref{ablation_F_int}c). These figures clearly highlight the effect of Fourier features. In particular, increasing the Fourier feature parameter enhances the resolution of the embedded inhomogeneity profile.\\
The next component of the proposed method we consider is the architecture of the NODE. In this regard, we define three cases: a coarse architecture [1, 2, 1], a finer one [1, 4, 4, 4, 1], and the finest [1, 8, 8, 8, 1]. The corresponding results are shown in Fig \ref{ablation_NODE}b, Fig \ref{ablation_NODE}c, and Fig \ref{ablation_NODE}d, respectively, with the ground truth data shown in Fig \ref{ablation_NODE}a. As can be seen, the architecture [1, 4, 4, 4, 1] provides a better identification of the inhomogeneity profile, slightly outperforming [1, 8, 8, 8, 1]. This suggests that over-parameterization of the NODE may have an adverse effect on the performance of the inverse problem.\\
Finally, we investigate the effect of different hyper-network architectures on material identification using the proposed method. To this end, we consider three configurations: a coarse architecture [20, 10, 126] with Fourier features [2, 10], a finer architecture [80, 40, 40, 126] with Fourier features [2, 40], and the finest [80, 60, 60, 126] with Fourier features [2, 80]. The corresponding results are shown in Fig \ref{ablation_hyper}b, Fig  \ref{ablation_hyper}c, and Fig \ref{ablation_hyper}d, respectively. Similar to the NODE architecture, an over-parameterization effect is observed in the identification of the inhomogeneity profile when using the most complex hyper-network architecture. This observation suggests that beyond a certain complexity, additional network depth may not yield significant improvements in inhomogeneity profile identification.
\subsection{Method Breakdown}
\begin{figure}[h!]
    \centering
    % \subfile{figures/fig_architecture}
        \begin{adjustbox}{center}
            \includegraphics[width=1.1\textwidth]{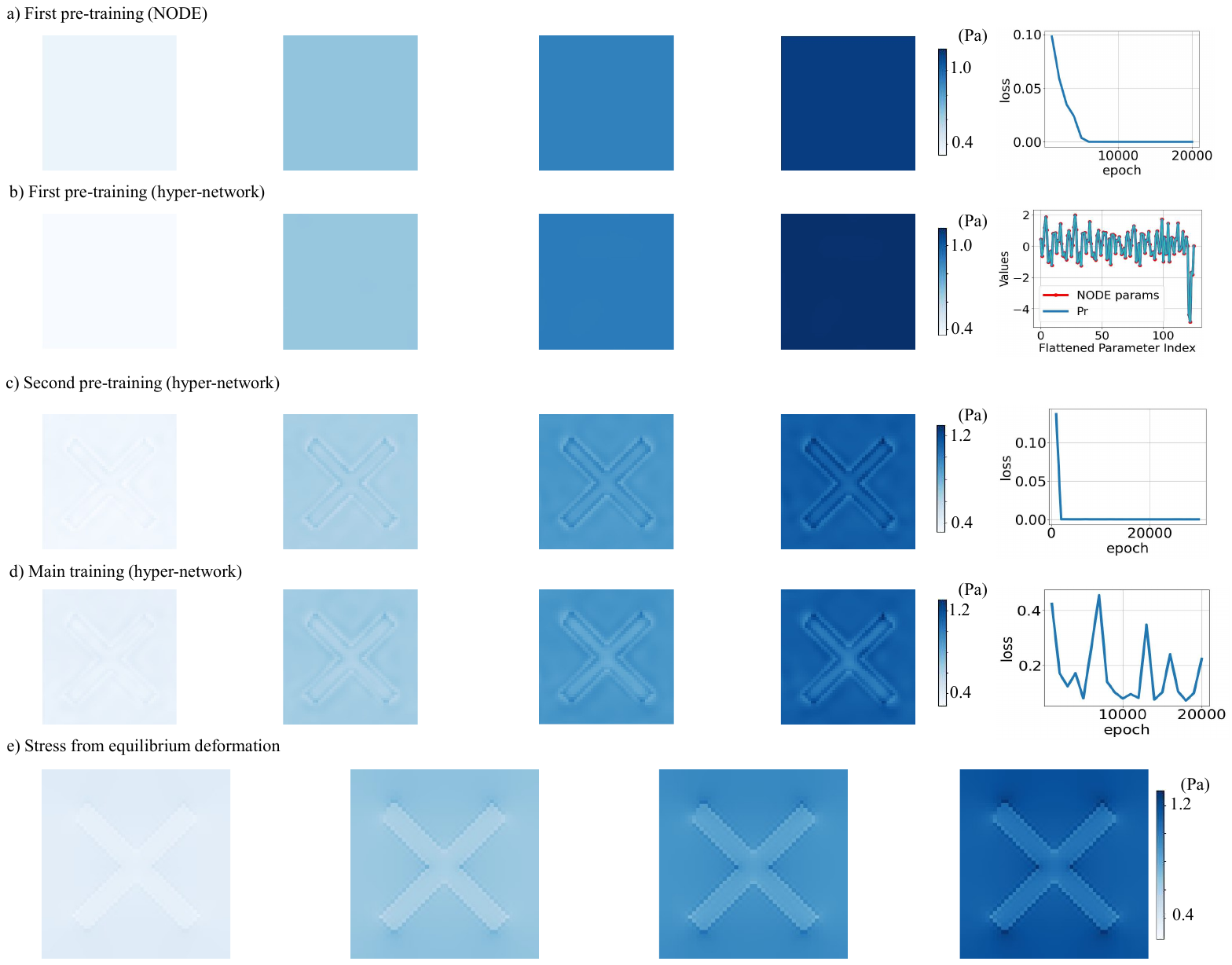}
    \end{adjustbox}
    \caption{An illustrative example showing the results of the major steps in the proposed inverse method. Panel (a) presents the results after training a homogeneous NODE to recover the average stress field. Panel (b) shows the results after training the hyper-network to generate the parameters of the homogeneous NODE. The stress field resulting from a non-homogeneous deformation gradient after the second pre-training is shown in panel (c). Panel (d) presents the final result after the second pre-training. The ground truth field corresponding to panel (d) is also shown for comparison in panel (e). }
    \label{detailed_results} 
\end{figure}

We present detailed results of the proposed method for the example with an embedded X-shaped inhomogeneity, as illustrated in Fig \ref{detailed_results}. Fig \ref{detailed_results}a shows the outcome of the first training phase, where a homogenized field is learned. Fig  \ref{detailed_results}b displays the homogenized field after training the hyper-network to reproduce the NODE parameters. As shown, the hyper-network successfully captures the NODE parameters and accurately reconstructs the homogenized field. Subsequently, Fig \ref{detailed_results}c presents the results of the second pre-training step, in which the hyper-network is trained to define a NODE field that yields average stresses corresponding to a non-homogeneous strain field. Finally, the stress field obtained from the final training phase is shown in Fig.\ref{detailed_results}d, alongside the ground truth equilibrium stress field in Fig. \ref{detailed_results}e for validation. As observed, the proposed method yields results in good agreement with the ground truth.    

%% References
\clearpage
 \bibliographystyle{elsarticle-num} 
 \bibliography{references}
\end{document}